\pdfoutput=1

\documentclass[11pt]{article}

\usepackage[review]{acl}

\usepackage{times}
\usepackage{latexsym}

\usepackage[T1]{fontenc}

\usepackage[utf8]{inputenc}

\usepackage{kotex}
\usepackage{multirow}
\usepackage{graphicx}
\usepackage{bm}
\usepackage{amsmath}
\usepackage{amssymb}

\usepackage{microtype}

\title{CoMPM: Context Modeling with Speaker's Pre-trained Memory Tracking for Emotion Recognition in Conversation}

\author{Joosung Lee, Wooin Lee \\
  Kakao Enterprise Corp., South Korea \\
  \texttt{\{rung.joo, dan.kes\}@kakaoenterprise.com} \\}

\begin{document}
\maketitle
\begin{abstract}
As the use of interactive machines grow, the task of Emotion Recognition in Conversation (ERC) became more important. If the machine-generated sentences reflect emotion, more human-like sympathetic conversations are possible. Since emotion recognition in conversation is inaccurate if the previous utterances are not taken into account, many studies reflect the dialogue context to improve the performances. Many recent approaches show performance improvement by combining knowledge into modules learned from external structured data. However, structured data is difficult to access in non-English languages, making it difficult to extend to other languages. Therefore, we extract the pre-trained memory using the pre-trained language model as an extractor of external knowledge. We introduce CoMPM, which combines the speaker's pre-trained memory with the context model, and find that the pre-trained memory significantly improves the performance of the context model. CoMPM achieves the first or second performance on all data and is state-of-the-art among systems that do not leverage structured data. In addition, our method shows that it can be extended to other languages because structured knowledge is not required, unlike previous methods. Our code is available on github~\footnote{https://github.com/rungjoo/CoMPM}.
\end{abstract}

\section{Introduction}
As the number of applications such as interactive chatbots or social media that are used by many users has recently increased dramatically, Emotion Recognition in Conversation (ERC) plays a more important role in natural language processing, and as a proof, a lot of research~\citep{ERC-research, ijcai2019-752, ghosal-etal-2020-cosmic, JiaoLK20} has been conducted on the task. 

The ERC module increases the quality of empathetic conversations with the users and can be utilized when sending tailored push messages to the users~\citep{shin2019happybot, ZandieM20, Lin_Xu_2020}. In addition, emotion recognition can be effectively used for opinion mining, recommender systems, and healthcare systems where it can improve the service qualities by providing personalized results. As these interactive machines increase, the ERC module plays an increasingly important role.

\begin{figure}[!t]
    \centering 
    \includegraphics[width=0.9\columnwidth]{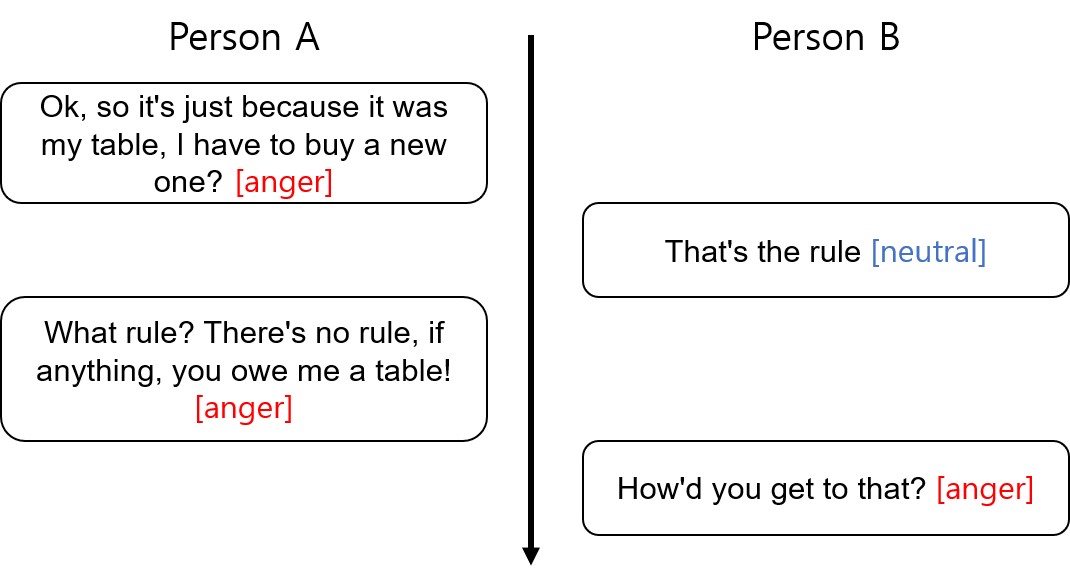}
    \caption{An example of MELD dataset}
    \label{fig:dialog}
\end{figure}

Figure~\ref{fig:dialog} is an example of a conversation in which two speakers are angry at each other. The emotion of speaker B's utterance (\textit{"How'd you get to that?"}) is \textit{angry}. If the system does not take into account previous utterances, it is difficult to properly recognize emotions. Like the previous studies~\cite{ghosal-etal-2020-cosmic}, we show that the utterance-level emotion recognition, which does not consider the previous utterance, have limitations and experiments result in poor performances. 

Therefore, recent studies are attempting to recognize emotions while taking into account the previous utterances. Representatively, DialogueRNN~\cite{DialogueRNN} recognizes the present emotion by tracking context from the previous utterances and the speaker's emotion. AGHMN~\cite{JiaoLK20} considers the previous utterances through memory summarizing using GRU with attention.

Many recent studies use external knowledge to improve the ERC performance. However, this external knowledge is often only available in English. In order to utilize the previous methods in languages of other countries, it is expensive and difficult to utilize because external knowledge data must be newly constructed. In recent NLP studies, due to the effectiveness of the pre-trained language model, it has already been developed in many countries. Since pre-trained language models are trained by unsupervised learning, these models are relatively usable approaches regardless of language types. \citet{petroni-etal-2019-language} introduces that these language models can be used as knowledge bases and have many advantages over the structured knowledge bases. Based on these studies, we eliminate the dependence on structured external data used in cutting-edge systems and use a pre-trained language model as a feature extractor of knowledge.

CoMPM, introduced in this paper, is composed of two modules that take into account previous utterances in dialogue. (1) The first is a context embedding module (CoM) that reflects all previous utterances as context. CoM is an auto-regressive model that predicts the current emotion through attention between the previous utterances of the conversation and the current utterance. (2) The second is a pre-trained memory module (PM) that extracts memory from utterances. We use the output of the pre-trained language model as the memory embedding where the utterances are passed into the language model. We use the PM to help predict the emotion of the speaker by taking into account the speaker's linguistic preferences and characteristics.

We experiment on 4 different English ERC datasets. Multi-party datasets are MELD~\cite{poria-etal-2019-meld} and EmoryNLP~\cite{emorynlp}, and dyadic datasets are IEMOCAP~\cite{iemocap} and DailyDialog~\cite{dailydialog}. CoMPM achieves the first or second performance according to the evaluation metric compared to all previous systems. We perform an ablation study on each module to show that the proposed approach is effective. Further experiments also show that our approach can be used in other languages and show the performance of CoMPM when the number of data is limited.

\section{Related Work}
Many recent studies use external knowledge to improve the ERC performance. KET~\cite{zhong-etal-2019-knowledge} is used as external knowledge based on ConceptNet~\cite{conceptnet} and emotion lexicon NRC\_VAD~\cite{mohammad-2018-obtaining} as the commonsense knowledge. ConceptNet is a knowledge graph that connects words and phrases in natural language using labeled edges. NRC\_VAD Lexicon has human ratings of valence, arousal, and dominance for more than 20,000 English words. COSMIC~\cite{ghosal-etal-2020-cosmic} and Psychological~\cite{li-etal-2021-past-present} improve the performance of emotion recognition by extracting commonsense knowledge of the previous utterances. Commonsense knowledge feature is extracted and leveraged with COMET~\cite{bosselut-etal-2019-comet} trained with ATOMIC (The Atlas of Machine Commonsense)~\cite{ATOMIC}. ATOMIC has 9 sentence relation types with inferential if-then commonsense knowledge expressed in text. ToDKAT~\cite{zhu-etal-2021-topic} improves performance by combining commonsense knowledge using COMET and topic discovery using VHRED~\cite{Serban} to the model. 

Ekman~\cite{Ekman1992AnAF} constructs taxonomy of six common emotions (Joy, Sadness, Fear, Anger, Surprise, and Disgust) from human facial expressions. In addition, Ekman explains that a multi-modal view is important for multiple emotions recognition. The multi-modal data such as MELD and IEMOCAP are some of the available standard datasets for emotion recognition and they are composed of text, speech and vision-based data. \citet{audiovisual} uses speech and visual information to recognize emotions, and \cite{alm-etal-2005-emotions} attempts to recognize emotions based on text information. MELD and ICON~\cite{hazarika-etal-2018-icon} show that the more multi-modal information is used, the better the performance and the text information plays the most important role. Multi-modal information is not always given in most social media, especially in chatbot systems where they are mainly composed of text-based systems. In this work, we design and introduce a text-based emotion recognition system using neural networks.

In the previous studies, such as ~\citet{hazarika-etal-2018-conversational, zadeh-etal-2017-tensor, DialogueRNN}, most works focused on dyadic-party conversation. However, as the multi-party conversation datasets including MELD and EmoryNLP have become available, a lot of recent research is being conducted on multi-party dialogues such as~\citet{ijcai2019-752, JiaoLK20, ghosal-etal-2020-cosmic}. In general, the multi-party conversations have higher speaker dependency than the dyadic-party dialogues, therefore have more conditions to consider and result in poor performance.

\citet{ijcai2018-643, ijcai2018-639} shows that commonsense knowledge is important for understanding conversations and generating appropriate responses. ~\citet{liu2020towards} reports that the lack of external knowledge makes it difficult to classify implicit emotions from the conversation history. EDA~\cite{bothe-etal-2020-eda} expands the multi-modal emotion datasets by extracting dialog acts from MELD and IEMOCAP and finds out that there is a correlation between dialogue acts and emotion labels.

\section{Approach}

\subsection{Problem Statement}
\label{problem-Statement}
In a conversation, $M$ sequential utterances are given as $[(u_1, p_{u_1}), (u_2, p_{u_2}), ..., (u_M, p_{u_M})]$. $u_i$ is the utterance which the speaker $p_{u_i}$ uttered, where $p_{u_i}$ is one of the conversation participants. While $p_{u_i}$ and $p_{u_j}$ ($i\neq j$) can be the same speaker, the minimum number of the unique conversation participants should be 2 or more. The ERC is a task of predicting the emotion $e_t$ of $u_t$, the utterance of the $t$-th turn, given the previous utterances $h_t=\{u_1, ..., u_{t-1}\}$. Emotions are labeled as one of the predefined classes depending on the dataset, and the emotions we experimented with are either 6 or 7. We also experimented with a sentiment classification dataset which provides sentiment labels consisting of positive, negative and neutral.

\subsection{Model Overview}
\label{model-overivew}
\begin{figure*}[!t]
    \centering 
    \includegraphics[width=1.8\columnwidth]{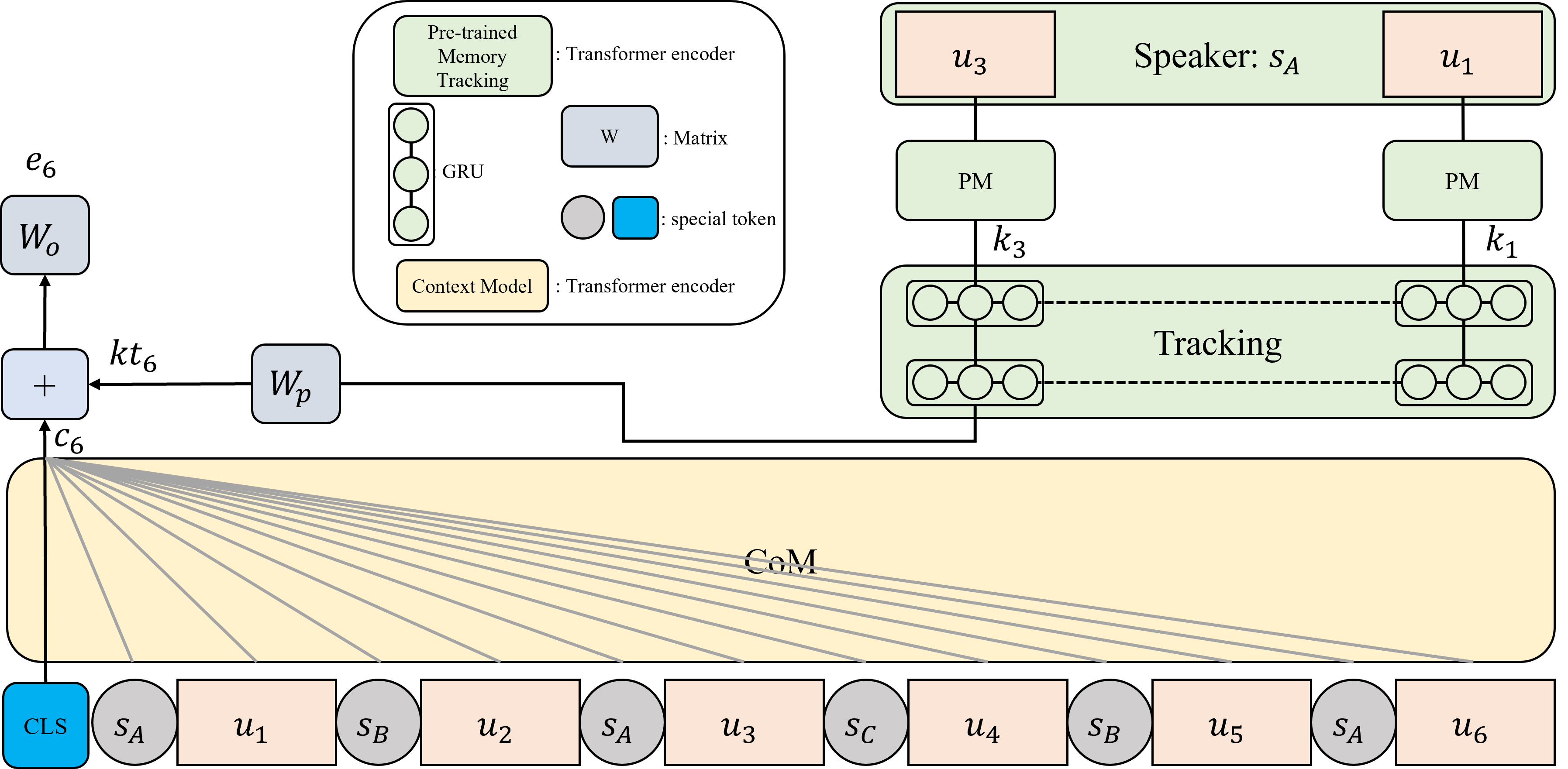}
    \caption{Our model consists of two modules: a context embedding module and a pre-trained memory module. The figure shows an example of predicting emotion of $u_6$, from a 6-turn dialogue context. A, B, and C refer to the participants in the conversation, where $s_A=p_{u_1}=p_{u_3}=p_{u_6}, s_B=p_{u_2}=p_{u_5}, s_C=p_{u_3}$. $\mathbf{W_o}$ and $\mathbf{W_p}$ are linear matrices.}
    \label{fig:Model}
\end{figure*}

Figure~\ref{fig:Model} shows an overview of our model. Our ERC neural network model is composed of two modules. The first is CoM which catches the underlying effect of all previous utterances on the current speaker's emotions. Therefore, we propose a context model to handle the relationship between the current and the previous utterances. The second one is PM that leverages only the speaker's previous utterances, through which we want to reflect the speaker's knowledge.

If the CoM and PM are based on different backbones, we consider them to be unaligned with respect to each other's output representations. Therefore, we design the PM to follow CoM so that the output representations of CoM and PM can mutually understand each other. If CoM and PM are based on different architectures, CoMPM is trained to understand each other's representations by matching dimensions using $\mathbf{W_p}$ in Equation~\ref{eq:tracking_equ2}. The combination of CoM and PM is described in Section~\ref{sec:change_backbone}.

\subsection{CoM: Context Embedding Module}
The context embedding module predicts $e_t$ by considering all of the utterances before the $t$-th turn as the dialogue context. The example in Figure~\ref{fig:Model} shows how the model predicts the emotion of $u_6$ uttered by $s_A$, given a conversation of three participants ($s_A$, $s_B$, $s_C$). The previous utterances are $h_6 = \{u_1, \cdots u_5 \}$ and $e_6$ is predicted while considering the relationship between $u_6$ and $h_6$.

We consider multi-party conversations where 2 or more speakers are involved. A special token $<$$s_{\mathbb{P}}$$>$ is introduced to distinguish participants in the conversation and to handle the speaker's dependency where $\mathbb{P}$ is the set of participants. In other words, the same special token appears before the utterances of the same speaker.

We use an Transformer encoder as a context model. In many natural language processing tasks, the effectiveness of the pre-trained language model has been proven, and we also set the initial state of the model to RoBERTa~\cite{liu2019roberta}. RoBERTa is an unsupervised pre-trained model with large-scale open-domain corpora of unlabeled text. 

We use the embedding of the special token $<$cls$>$ to predict emotion. The $<$cls$>$ token is concatenated at the beginning of the input and the output of the context model is as follows:
\begin{equation}
\label{eq:COM_equ}
\bm{c}_{t} = \textrm{CoM} (<\textrm{cls}>, \mathbb{P}_{:t-1}, h_t, u_t)
\end{equation}
where $\mathbb{P}_{:t-1}$ is the set of speakers in the previous turns. $\bm{c}_{t} \in \mathbb{R}^{1 \times h_c}$ and $h_c$ is the dimension of CoM.

\subsection{PM: Pre-trained Memory Module}
External knowledge is known to play an important role in understanding conversation. Pre-trained language models can be trained on numerous corpora and be used as an external knowledge base. Inspired by previous studies that the speaker's knowledge helps to judge emotions, we extract and track pre-trained memory from the speaker's previous utterances to utilize the emotions of the current utterance $u_t$. If the speaker has never appeared before the current turn, the result of the pre-trained memory is considered a zero vector.

Since $<$cls$>$ is mostly used for the task of classifying sentences, we use the embedding output of the $<$cls$>$ token as a vector representing the utterance as follows:

\begin{equation}
\label{eq:CK_equ}
\bm{k}_{i} = \textrm{PM} (<\textrm{cls}>, u_i)
\end{equation}
where $p_{u_i} = p_{S}$, S is the speaker of the current utterance. $\bm{k}_{i} \in \mathbb{R}^{1 \times h_{k}}$ and $h_k$ is the dimension of PM.

\subsection{CoMPM: Combination of CoM and PM}
We combine CoM and PM to predict the speaker's emotion. In many dialogue systems~\cite{zhang-etal-2018-modeling, ma-etal-2019-triplenet}, it is known that utterances close to the current turn are important for response. Therefore, we assume that utterances close to the current utterance will be important in emotional recognition. 

\subsubsection{Tracking Method}
\label{sec:tracking}
We use $\bm{k}_{i}$ tracking method using GRU. The tracking method assumes that the importance of all previous speaker utterances to the current emotion is not equal and varies with the distance of the current utterance. In other words, since the flow of conversation changes as it progresses, the effect on emotion may differ depending on the distance from the current utterance. We track and capture the sequential position information of $\bm{k}_{i}$ using a unidirectional GRU:

\begin{equation}
\label{eq:tracking_equ}
\bm{kt}_{t} = \textrm{GRU} (\bm{k}_{i_1}, \bm{k}_{i_2}, ..., \bm{k}_{i_n})
\end{equation}
where $t$ is the turn index of the current utterance, $n$ is the number of previous utterances of the speaker, and $i_s$ ($s=1,2,...,n$) is each turn uttered. $\bm{kt}_{t} \in \mathbb{R}^{1 \times h_{c}}$ is the output of $\bm{k}_{i_n}$ and as a result, the knowledge of distant utterance is diluted and the effect on the current utterance is reduced.

GRU is composed of 2-layers, the dimension of the output vector is $h_{c}$, and the dropout is set to 0.3 during training. Finally, the output vector $\bm{o}_{t}$ is obtained by adding $\bm{kt}_{t}$ and $\bm{c}_{t}$ in Equation~\ref{eq:tracking_equ2}.

\begin{equation}
\label{eq:tracking_equ2}
\bm{o}_{t} = \bm{c}_{t} + \mathbf{W_p}(\bm{kt}_{t})
\end{equation}

where, $\mathbf{W_p}$ is a matrix that projects the pre-trained memory to the dimension of the context output, and is used only when PM and CoM are different pre-trained language models.

\subsubsection{Emotion Prediction}
Softmax is applied to the vector multiplied by $\bm{o}_{t}$ and the linear matrix $\mathbf{W_o} \in \mathbb{R}^{h_e \times h_c}$ to obtain the probability distribution of emotion classes, where $h_e$ is the number of emotion classes. $e_t$ is the predicted emotion class that corresponds to the index of the largest probability from the emotion class distribution.
\begin{equation}
\label{eq:final_equ}
P(e) = \textrm{softmax}(\mathbf{W_o}(\bm{o}_{t}))
\end{equation}
The objective is to minimize the cross entropy loss so that $e_t$ is the same as the ground truth emotional label.

\begin{table*}[!t]
\centering
\resizebox{1.8\columnwidth}{!}{
\begin{tabular}{|c|c|c|c|c|c|c|c|c|}
\hline
\multirow{2}{*}{Dataset} & \multicolumn{3}{c|}{dialogues} & \multicolumn{3}{c|}{utterance} & \multirow{2}{*}{classes} & \multirow{2}{*}{Evaluation Metrics} \\ \cline{2-7}
                         & train     & dev      & test    & train     & dev      & test    &                          &                                     \\ \hline\hline
IEMOCAP                  & 108       & 12       & 31      & 5163      & 647      & 1623    & 6                        & weighted avg F1                     \\ \hline
DailyDialog              & 11118     & 1000     & 1000    & 87170     & 8069     & 7740    & 7(6)                     & Macro F1 \& Micro F1                   \\ \hline
MELD                     & 1038      & 114      & 280     & 9989      & 1109     & 2610    & 3, 7                     & weighted avg F1                     \\ \hline
EmoryNLP                 & 713       & 99       & 85      & 9934      & 1344     & 1328    & 3, 7                     & weighted avg F1                     \\ \hline
\end{tabular}
}
\caption{Statistics and descriptions for the four datasets. DailyDialog uses 7 classes for training, but we measure Macro-F1 for only 6 classes excluding neutral. MELD and EmoryNLP are used to measure weighted avg F1 for both emotion (7) and sentiment (3) classes.}
\label{Tab:dataset}
\end{table*}

\section{Experiments}
\subsection{Dataset}
We experiment on four benchmark datasets. MELD~\cite{poria-etal-2019-meld} and EmoryNLP~\cite{emorynlp} are multi-party datasets, while IEMOCAP~\cite{iemocap} and DailyDialog~\cite{dailydialog} are dyadic-party datasets. The statistics of the dataset are shown in Table~\ref{Tab:dataset}.

IEMOCAP is a dataset involving 10 speakers, and each conversation involves 2 speakers and the emotion-inventory is given as "happy, sad, angry, excited, frustrated and neutral". The train and development dataset is a conversation involving the previous eight speakers, and the train and development are divided into random splits at a ratio of 9:1. The test dataset is a conversation involving two later speakers.

DailyDialog is a dataset of daily conversations between two speakers and the emotion-inventory is given as "anger, disgust, fear, joy, surprise, sadness and neutral". Since more than 82\% of the data are tagged as neutral, neutral emotions are excluded when evaluating systems with Micro-F1 as did in the previous studies.

MELD is a dataset based on Friends TV show and provides two taxonomy: emotion and sentiment. MELD's emotion-inventory is given as "anger, disgust, sadness, joy, surprise, fear and neutrality" following Ekman~\cite{Ekman1992AnAF} and sentiment-inventory is given as "positive, negative and neutral".

EmoryNLP, like MELD, is also a dataset based on Friends TV show, but the emotion-inventory is given as "joyful, peaceful, powerful, scared, mad, sad and neutral". Sentiment labels are not provided, but sentiment classes can be grouped as follows: positive: \{joyful, peaceful, powerful\}, negative: \{scared, mad, sad\}, neutral: \{neutral\}

\subsection{Training Setup}
We use the pre-trained model from the huggingface library~\footnote{https://github.com/huggingface/transformers}. The optimizer is AdamW and the learning rate is 1e-5 as an initial value. The learning rate scheduler used for training is \textit{get\_linear\_schedule\_with\_warmup}, and the maximum value of 10 is used for the gradient clipping. We select the model with the best performance on the validation set. All experiments are conducted on one V100 GPU with 32GB memory.

\begin{table*}[!t]
\centering
\resizebox{2.0\columnwidth}{!}{
\begin{tabular}{c|c|cc|cc|cc}
\hline
\multirow{2}{*}{Models} & IEMOCAP        & \multicolumn{2}{c|}{DailyDialog} & \multicolumn{2}{c|}{MELD}           & \multicolumn{2}{c}{EmoryNLP}       \\ \cline{2-8} 
                        & W-Avg F1       & Macro F1        & Micro F1       & W-Avg F1 (3-cls) & W-Avg F1 (7-cls) & W-Avg F1 (3-cls) & W-Avg F1 (7-cls) \\ \hline\hline
KET*                     & 59.56          & -               & 53.37          & -                & 58.18            & -                & 34.39            \\
RoBERTa   DialougeRNN   & 64.76          & 49.65           & 57.32          & 72.14            & 63.61            & 55.36            & 37.44            \\
RGAT+P                  & 65.22          & -               & 54.31          & -                & 60.91            & -                & 34.42            \\
HiTrans                 & 64.5           & -               & -              & -                & 61.94            & -                & 36.75            \\
COSMIC*                 & 65.28          & 51.05           & 58.48          & \textbf{73.2}    & 65.21            & \textbf{56.51}   & 38.11            \\
ERMC-DisGCN             & -              & -               & -     & -                & 64.22            & -                & 36.38            \\
Psychological*           & \textbf{66.96}          & 51.95           & \textbf{59.75} & -                & 65.18            & -                & 38.8             \\
DialogueCRN           & 66.05          & -           & - & -                & 58.39            & -                & -             \\
ToDKAT*                 & 61.33          & \textbf{52.56}  & 58.47          & -                & \textbf{65.47}   & -                & \textbf{43.12}   \\ \hline\hline
CoMPM                   & 66.33          & \textbf{53.15}  & \textbf{60.34} & \textbf{73.08}   & \textbf{66.52}   & \textbf{57.14}   & 37.37            \\
CoM                     & 65.05          & 51.17           & 58.63          & 71.67            & 64.9             & 56.27            & 36.34            \\
PM                      & 52.56          & 49.08           & 56.23          & 69.21            & 63.4             & 53.87            & 35.48            \\
CoMPM(f)                 & \textbf{69.46} & 51.67      & 59.02          & 73.04            & 65.77            & 55.44            & \textbf{38.93}   \\ 
CoMPM(s)                 & 64.68 & 48.86      & 55.81          & 71.97            & 
65.12            & 53.66            & 34.72   \\
CoMPM(k)                 & 64.3 & 52.33      & 59.09          & 72.67            & 66.22            & 56.62            & 36.96   \\ \hline
\end{tabular}
}
\caption{Comparison of our models with various previous models and the results on 4 datasets. Our models are trained 3 times for each experiment and the average of the scores is evaluated (same in other tables). Test performance is measured by the model with the best score in the validation dataset. Bold text indicates the best performance in each part (comparative models or ours). * indicates models that leverages structured external data.}
\label{Tab:results}
\end{table*}

\subsection{Previous Method}
We show that the proposed approach is effective by comparing it with various baselines and the state-of-the-art methods.

\textbf{KET}~\cite{zhong-etal-2019-knowledge} is a Knowledge Enriched Transformer that reflects contextual utterances with a hierarchical self-attention and leverages external commonsense knowledge by using a context-aware affective graph attention mechanism.

\textbf{DialogueRNN}~\cite{DialogueRNN} uses a GRU network to keep track of the individual party states in the conversation to predict emotions. This model assumes that there are three factors in emotion prediction: the speaker, the context from the preceding utterances and the emotion of the preceding utterances. Also, ~\citet{ghosal-etal-2020-cosmic} shows the performance of \textbf{RoBERTa+DialogueRNN} when the vectors of the tokens are extracted with a pre-trained RoBERTa.

\textbf{RGAT+P}~\cite{ishiwatari-etal-2020-relation} (relational graph attention networks) proposes relational position encodings with sequential information reflecting the relational graph structure, which shows that both the speaker dependency and the sequential information can be captured.

\textbf{HiTrans}~\cite{li-etal-2020-hitrans} proposes a transformer-based context- and speaker-sensitive model. HiTrans utilize BERT as the low-level transformer to generate local utterance representations, and feed them into another high-level transformer.

\textbf{COSMIC}~\cite{ghosal-etal-2020-cosmic} incorporates different elements of commonsense such as mental states, events and causal relations, and learns the relations between participants in the conversation. This model uses pre-trained RoBERTa as a feature extractor and leverages COMET trained with ATOMIC as the commonsense knowledge.

\textbf{ERMC-DisGCN}~\cite{sun-etal-2021-discourse-aware} proposes a discourse-aware graph neural network that utilizes self-speaker dependency of interlocutors as a relational convolution and informative cues of dependent utterances as a gated convolution.

\textbf{Psychological}~\cite{li-etal-2021-past-present} uses commonsense knowledge as enrich edges and processes it with graph transformer. Psychological performs emotion recognition by utilizing intention of utterances from not only past contexts but also future context.

\textbf{DialogueCRN}~\cite{hu-etal-2021-dialoguecrn} introduces an intuitive retrieving process, the reasoning module, which understands both situation-level and speaker-level contexts.

\textbf{ToDKAT}~\cite{zhu-etal-2021-topic} proposes a language model with topic detection added, and improves performance by combining it with commonsense knowledge.  The performance of ToDKAT in MELD was re-released on github~\footnote{https://github.com/something678/TodKat}.

\subsection{Result and Analysis}
Table~\ref{Tab:results} shows the performance of the previous methods and our models. CoM used alone does not leverage PM and predicts emotions by considering only the dialogue context. PM used alone is not used as a memory module, but the same backbone is used. PM used alone predicts emotion only with the utterance of the current turn without considering the context. CoMPM is a model in which both CoM and PM parameters are updated in the initial state of the pre-trained LM. CoMPM(f) is a model in which PM parameters are frozen in the initial state (pre-trained LM) and is not trained further, and CoMPM(s) is a model in which PM is trained from scratch. CoMPM(k) is a model in which PM is trained on ConceptNet. Following previous studies, we use the average vector for each token in PM(k) as the feature of the utterance. We use the pre-trained model provided by the site~\footnote{https://huggingface.co/HungChau/distilbert-base-uncased-concept-extraction-kp20k-v1.2-concept-extraction-allwikipedia-v1.0} as PM(k).

The effect of PM can be confirmed through the performance comparison between CoM and CoMPM, and the effect of CoM can be confirmed by comparing the results of CoM and PM. Since PM does not consider context, it showed worse performance than CoM, and the performance gap is larger in the IEMOCAP dataset with a higher average number of conversation turns. As a result, we show that the combination of CoM and PM is effective in achieving better performance.

We confirm the effect of PM structure in the model through the performance of CoMPM(s). If PM parameters are not frozen and are instead randomly initialized (i.e. PM(s)), the performance deteriorates. CoMPM(s) performs worse than CoMPM, and even performs worse than CoM on the other datasets except for MELD. That is, PM(s) cannot be regarded as a pre-trained memory because the parameters are randomly initialized, and simply increasing the model complexity does not help to improve the performance. CoMPM(f) shows similar performance to CoMPM and achieves better performance depending on the data. PM(f) is not fine-tuned on the data, but it extracts general pre-trained memory from a pre-trained language model. The comparison between PM and PM(f) will be further described in Section~\ref{sec:less}. In addition, CoMPM(k) shows better performance than CoM, PM, and CoMPM(s) except for IEMOCAP. In IEMOCAP, CoMPM(k) has lower performance than CoM. For all datasets, CoMPM(k) performs slightly worse than CoMPM. In other words, ConceptNet improves the performance of CoMPM, but is not as effective as pre-trained memory. As a result, we regard pre-trained memory as compressed knowledge, which can play a role similar to external knowledge used in cutting-edge systems.


The best performance of our approaches is CoMPM or CoMPM(f), both of which combine pre-trained memory. We achieve state-of-the-art performance among all systems that do not leverage structured external data and achieve the first or second performance even including systems that leverage external data. Therefore, our approach can be extended to other languages without structured external data as well, which is described in Section~\ref{sec:other_languages}.

\begin{table*}[!t]
\centering
\resizebox{1.6\columnwidth}{!}{
\begin{tabular}{c|c|c|cc|c|c}
\hline
\multirow{2}{*}{CoM} & \multirow{2}{*}{PM(f)} & IEMOCAP                                                      & \multicolumn{2}{c|}{DailyDialog}                                                                              & MELD                                                         & EmoryNLP                                              \\ \cline{3-7} 
                     &                        & W-Avg F1                                                     & Macro F1                                              & Micro F1                                              & W-Avg F1 (7-cls)                                             & W-Avg F1 (7-cls)                                      \\ \hline\hline
RoBERTa              & BERT                   & \begin{tabular}[c]{@{}c@{}}65.93\\      (-3.53)\end{tabular} & \begin{tabular}[c]{@{}c@{}}52.74\\      (+1.07)\end{tabular} & \begin{tabular}[c]{@{}c@{}}59.97\\      (+0.95)\end{tabular} & \begin{tabular}[c]{@{}c@{}}65.41\\      (-0.36)\end{tabular} & \begin{tabular}[c]{@{}c@{}}37.25\\      (-1.68)\end{tabular}                                                 \\ \hline
RoBERTa              & GPT2                   & \begin{tabular}[c]{@{}c@{}}68.54\\      (-0.92)\end{tabular}        & \begin{tabular}[c]{@{}c@{}}50.68\\      (-0.99)\end{tabular} & \begin{tabular}[c]{@{}c@{}}59.61\\      (+0.59)\end{tabular} & \begin{tabular}[c]{@{}c@{}}65.58\\      (-0.19)\end{tabular}        & \begin{tabular}[c]{@{}c@{}}36.39\\      (-2.54)\end{tabular} \\ \hline
BERT                 & RoBERTa                & \begin{tabular}[c]{@{}c@{}}62.69\\      (-6.77)\end{tabular} & \begin{tabular}[c]{@{}c@{}}48.99\\      (-2.68)\end{tabular} & \begin{tabular}[c]{@{}c@{}}57.34\\      (-1.68)\end{tabular} & \begin{tabular}[c]{@{}c@{}}63.79\\      (-1.98)\end{tabular} & \begin{tabular}[c]{@{}c@{}}35.47\\      (-3.46)\end{tabular}                                                 \\ \hline
\end{tabular}
}
\caption{Emotion recognition performance according to the combination of different backbones of CoM and PM. The value in parentheses is the performance difference from the original CoMPM(f) (RoBERTa + RoBERTa). We use the bert-large-uncased and GPT2-medium versions.}
\label{Tab:CoMPM_variance}
\end{table*}

\subsection{Combinations of CoM and PM}
\label{sec:change_backbone}
We experiment with the effect of pre-trained memory of different language models. To eliminate the influence of the PM structure, we freeze the parameters of PM and use it as a feature extractor. Table~\ref{Tab:CoMPM_variance} shows the performance of the pre-trained memory extracted by the different language models. If PM and CoM are based on different backbones, the pre-trained memory is projected through $\mathbf{W_p}$ as the dimension of the context output. RoBERTa+BERT and RoBERTa+GPT2 (combination of CoM and PM(f)) have lower performance than RoBERTa+RoBERTa, which is inferred because pre-trained memory of RoBERTa contains richer information than BERT and GPT2. Since there is a lot of training data in the diallydialog and $\mathbf{W_p}$ is fine-tuned to the data to mutually understand the pre-trained memory and context representation. Therefore, we infer that performance does not decrease even if the PM changes from the dailydialog. However, even if other PMs are used, the performance is improved compared to using only CoM, so the pre-trained memory of other language models is also effective for emotion recognition.

BERT+RoBERTa has a larger performance decrease than RoBERTa+BERT. In particular, in IEMOCAP data with a long average number of turns in the context, the performance deteriorates significantly. In addition, the performance of BERT+RoBERTa is lower than CoM (RoBERTa), which supports that the performance of CoM is a more important factor than the use of pre-trained memory. In other words, we confirm that CoM is more important than PM in our system for performance, and it is effective to focus on context modeling rather than external knowledge in the study of emotion recognition in conversation.

\begin{figure}[!t]
    \centering 
    \includegraphics[width=1.0\columnwidth]{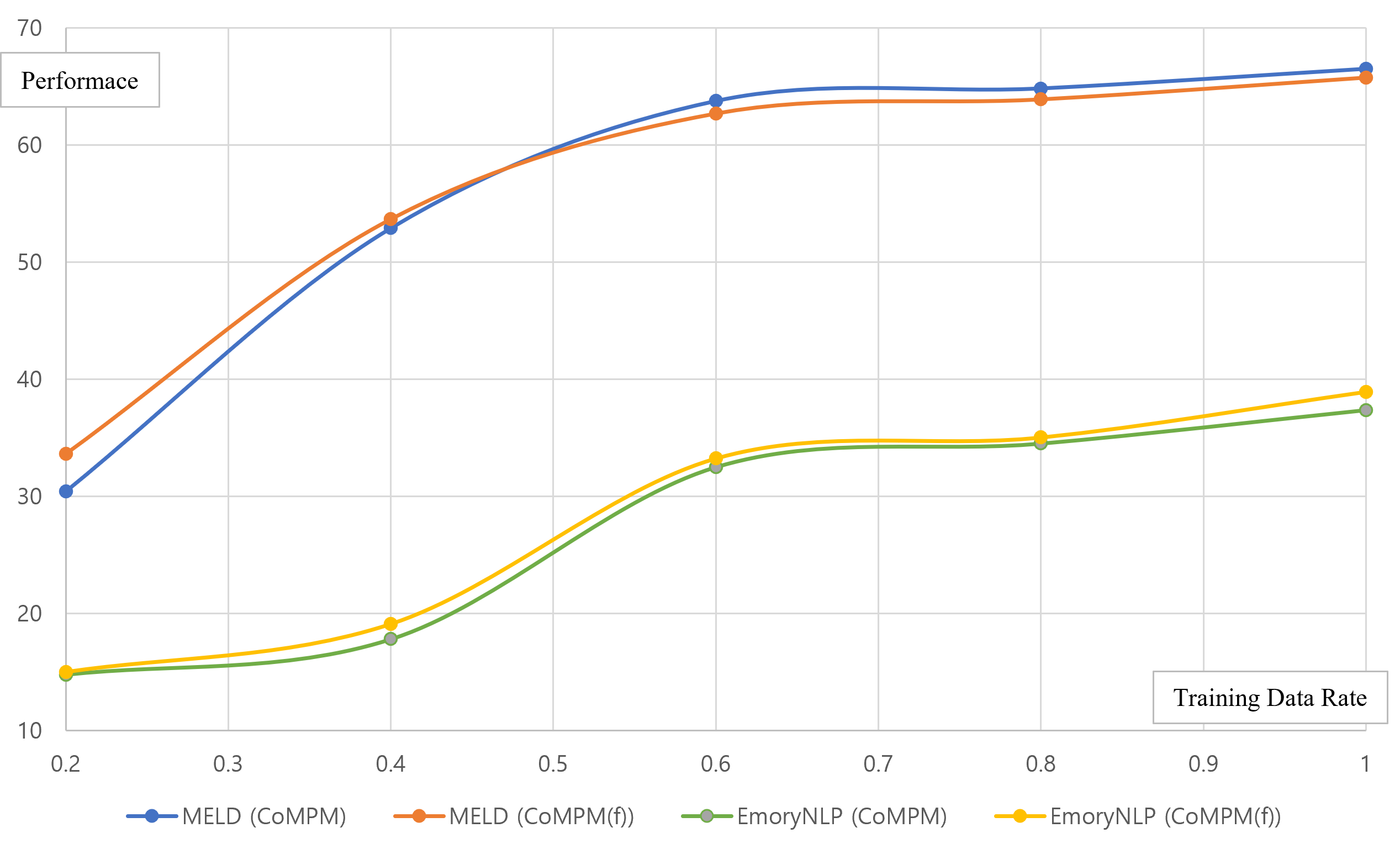}
    \caption{Performance according to the size of training data of MELD and EmoryNLP}
    \label{fig:training_rate}
\end{figure}

\subsection{Training with Less Data}
\label{sec:less}
CoMPM is an approach that eliminates dependence on external sources and is easily extensible to any language. However, the insufficient number of emotional data available in other countries (or actual service) remains a problem. Therefore, we conduct additional experiments according to the use ratio of training data in MELD and EmoryNLP, where there is neither too much nor too little data. Figure~\ref{fig:training_rate} shows the performance of the model according to the ratio of the training data. In MELD and EmoryNLP, even if only 60\% and 80\% are used, respectively, the performance decreases by only 3 points.

Table~\ref{Tab:results} shows that CoMPM(f) achieves better performance than CoMPM in the emotion classification of IMEOCAP and EmoryNLP, which has fewer training data than other settings. On the other hand, if there is a lot of training data, CoMPM shows better performance. Figure~\ref{fig:training_rate} shows that as the number of data decreases, CoMPM(f) shows better results than CoMPM, which indicates that it is better to freeze the parameters of PM when the number of training data is insufficient. Therefore, if there is a lot of training data in the real-world application, CoMPM is expected to achieve good performance, otherwise it is CoMPM(f).

\subsection{ERC in other languages}
\label{sec:other_languages}

Previous studies mostly utilize external knowledge to improve performance, but these approaches require additional publicly available data, which are mainly available for English. Indeed, structured knowledge and ERC data are lacking in other languages. Our approach can be extended to other languages without building additional external knowledge and achieves better performance than simply using a pre-trained model.

\subsubsection{Korean Dataset}
We constructed data composed of two speakers in Korean, and emotion-inventory is given as "surprise, fear, ambiguous, sad, disgust, joy, bored, embarrassed, neutral". The total number of sessions is 1000, and the average number of utterance turns is 13.4. We use the data randomly divided into train:dev:test in a ratio of 8:1:1. This dataset is for actual service and is not released to the public.

\subsubsection{Results in the Korean Dataset}

\begin{table}[!h]
\centering
\resizebox{0.45\columnwidth}{!}{
\begin{tabular}{|c|c|}
\hline
\multirow{2}{*}{Models} & Korean   \\ \cline{2-2} 
                        & W-Avg F1 \\ \hline\hline
PM                      & 31.86    \\ \hline
CoM                     & 57.46    \\ \hline
CoMPM                   & 60.66    \\ \hline
\end{tabular}
}
\caption{Results of our approaches in Korean.}
\label{Tab:Korean_results}
\end{table}

In Korean, our results are shown in Table~\ref{Tab:Korean_results}. The backbone of CoM and PM is Korean-BERT owned by the company, respectively. In the Korean dataset, like the English dataset, the performance is good in the order of CoMPM, CoM, and PM. Our approach simply shows a significant performance improvement on baselines that are fine-tuned to the language model and works well for other languages as well as for English.

\section{Conclusion}
We propose CoMPM that leverages pre-trained memory using a pre-trained language model. CoMPM consists of a context embedding module (CoM) and a pre-trained memory module (PM), and the experimental results show that each module is effective in improving the performance. CoMPM outperforms baselines on both dyadic-party and multi-party datasets and achieves state-of-the-art among systems that do not use external knowledge. In addition, CoMPM achieves performance comparable to cutting-edge systems that leverage structured external knowledge, which is the effect of pre-trained memory of the language model.

By combining other pre-trained memories, we find that the pre-trained memory extracted with RoBERTa is richer and more effective than the pre-trained memory extracted with BERT or GPT2. Since we believe that pre-trained memory is proportional to the performance of a language model, a language model with a large training corpus and many parameters is considered to be more effective. However, we find that context modeling is more important than pre-trained memory for emotion recognition in conversation, and future research will focus on context modeling.

Additionally, our approach achieves competitive performance and does not require externally structured data. Therefore, we show that it can be easily extended to Korean as well as English, and it is expected to be effective in other countries.

\bibliography{anthology,custom}
\bibliographystyle{acl_natbib}

\appendix

\end{document}